# LDCSF: Local depth convolution-based Swim framework for classifying multi-label histopathology images

Liangrui Pan
College of Computer Science and Electronic Engineering
HunanUniversity
Chang Sha, China
panlr@ hnu.edu.cn

Yutao Dou
College of Computer Science and Electronic Engineering
HunanUniversity
Chang Sha, China
ytdou@hnu.edu.cn

Zhichao Feng
Department of Radiology
Third Xiangya Hospital
Central South University
Chang Sha, China
fengzc2016@163.com

Liwen Xu*
College of Computer Science and Electronic Engineering
Hunan University
Chang Sha, China
xuliwen@hnu.edu.cn

Shaoliang Peng*
College of Computer Science and Electronic Engineering
Hunan University
Chang Sha, China
slpeng@hnu.edu.cn

*Abstract*—Histopathological images are the gold standard for diagnosing liver cancer. However, the accuracy of fully digital diagnosis in computational pathology needs to be improved. In this paper, in order to solve the problem of multi-label and low classification accuracy of histopathology images, we propose a locally deep convolutional Swim framework (LDCSF) to classify multi-label histopathology images. In order to be able to provide local field of view diagnostic results, we propose the LDCSF model, which consists of a Swin transformer module, a local depth convolution (LDC) module, a feature reconstruction (FR) module, and a ResNet module. The Swin transformer module reduces the amount of computation generated by the attention mechanism by limiting the attention to each window. The LDC then reconstructs the attention map and performs convolution operations in multiple channels, passing the resulting feature map to the next layer. The FR module uses the corresponding weight coefficient vectors obtained from the channels to dot product with the original feature map vector matrix to generate representative feature maps. Finally, the residual network undertakes the final classification task. As a result, the classification accuracy of LDCSF for interstitial area, necrosis, non-tumor and tumor reached 0.9460, 0.9960, 0.9808, 0.9847, respectively. Finally, we use the results of multi-label pathological image classification to calculate the tumor-to-stromal ratio, which lays the foundation for the analysis of the microenvironment of liver cancer histopathological images. Second, we released a multi-label histopathology image of liver cancer, our code and data are available at https://github.com/panliangrui/LSF.

## I. Introduction

Liver cancer is a common disease. According to data released by the World Health Organization (WHO), there were 410,000 new cases of liver cancer in China in 2020, accounting for 45% of the global incidence of liver cancer, including 390,000 deaths [1]. Currently, the five-year overall survival rate of liver cancer patients in China is only 14.1%. 70%-80% of patients are already in an unresectable state at initial diagnosis, and although many patients are diagnosed and treated in the early stages of liver cancer, its recurrence is still high. For patients in the middle and late stages of liver cancer, their prognosis is not optimistic; if cancer spreads to the surrounding lymph nodes, the 5-year survival rate of patients is only 11%. When cancer spreads to other organs, the 5-year survival rate is only 3%. Therefore, early diagnosis and screening of liver cancer are extremely important for the prognosis of liver cancer patients.

Histopathological images microscopically reflect the survival status of cancer cells and the degree of differentiation. Therefore, histological examination of histopathological sections is the gold standard for diagnosing and grading hepatocellular carcinoma [2]. Traditionally, pathologists use a microscope to observe histopathological images. With the development of digital image technology, pathologists can remotely view pathology images through a monitor to determine sensitive areas and malignancy. However, the pathologist's subjective judgment is sometimes influenced by experience, and localized views of the field of view often interfere with the whole slide diagnosis. To achieve an objective diagnosis, computational pathology is essential for quantitative image analysis. Classification and segmentation of tissue images is the first step in analyzing the microenvironment of liver cancer. For example, the classification task obtains relevant quantitative information such as the interstitial tumor ratio. In addition, the segmentation task obtains the distribution of sensitive regions of the tissue images. This helps to construct prognostic histological features of tumor subtypes and patient survival. Therefore, histological image classification techniques and segmentation techniques play an important role in computer-aided diagnosis and grading systems for cancer.

Recently, deep learning techniques have been widely used in the segmentation and classification of histopathology images and have demonstrated performance superior to existing methods. Nicolas et al. trained a deep convolutional neural network (Inception v3) on whole slide images (WSIs) obtained from The Cancer Genome Atlas to accurately and automatically classify them as LUAD, LUSC or normal lung tissue [3]. Chen et al. used deep learning for the classification and mutation prediction of

H&E images of liver cancer histopathology [4]. The deep learning model was trained using 2,123 pixel-level annotated WSIs of H&E staining and achieved near 100% sensitivity and 80.6% average specificity on a real-world test dataset [5]. A self-supervised convolutional neural network framework utilizes contextual, multi-resolution, and semantic features in pathology images for semi-supervised learning and domain adaptation and demonstrates the effectiveness of the Self-Path model on three different pathology datasets [6]. A multi-instance learning approach based on deep graph convolutional networks and feature selection (FS-GCN-MIL) can be used for histopathology image classification and predict lymph node metastasis [7]. Finally, the AI algorithm learns from 16 TB of data in the TCGA database through high-performance storage and GPU power. The results are evaluated by conservative "majority voting" to establish a subtype diagnostic consensus through vertical search and demonstrate high accuracy values for both frozen sections [8]. The widespread use of deep learning techniques has contributed to the development of histopathological image analysis.

We have to introduce several classical transformer-based networks. First, data-efficient image transformers (DeiT) is a dual intent and entity change architecture that has experimentally improved performance on complex multi-domain NLU datasets and achieved similar high performance on some simple datasets [9]. Second, tokens-to-token vision transformer (T2T-ViT) proposes a structured T2T module that can encode local information and defeats CNN-based models by careful design and does not require pre-training on giant training sets (e.g., JFT-300M) [10]. Pyramid vision transformer (PVT), unlike ViT, which usually has the low-resolution output and high computational and storage costs, PVT can not only be trained on dense partitions of images to achieve high output resolution but also can use progressively shrinking pyramids to reduce the computation of large feature maps [11]. Second, PVT inherits the advantages of CNN and Transformer by simply replacing the CNN backbone to make it a unified backbone in various vision tasks without convolution. The transformer in transformer (TNT) treats local patches (e.g., 16 × 16) as "visual sentences" and further divides them into smaller patches (e.g., four × four) as "visual words" [12]. The input data is encoded into powerful features by an attention mechanism. Cross-attention multi-scale vision transformer (CrossViT) uses a parallel framework combined with attention cross-fusion to extract features from the dataset and achieve better results on many publicly available datasets [13]. Finally, the Swin transformer contains sliding window (non-overlapping local window and overlapping cross-window) operations that can introduce the local nature of CNN convolution operations and, on the other hand, can save computational effort [14].

Nowadays, the attention-based mechanism of the transformer has been shown to outperform the best existing approaches in the field of computer vision [15][16]. Transformer-based multi-instance learning (MIL) explores the morphological and spatial information of histopathological images and efficiently handles balanced and unbalanced multi-classification tasks [17]. A novel MIL model based on the deformable transformer (DT) architecture and convolutional layers in embedded space can update each instance feature by globally aggregating the instance features in the package and encoding the location context information of the instance during package representation learning [18]. Self-supervised learning (SSL) model with a convolutional neural network (CNN) combined with a hybrid model designed with an improved transformer architecture (TransPath). a token aggregation and excitation (TAE) module is introduced in TransPath and placed after the self-attention of the converter encoder for capturing more global information [19]. Using the pathological relationship between primary tumors and their lymph node metastases, Zhihua et al. developed an effective attention-based mutual knowledge distillation (AMKD) paradigm [20]. The collected WSI dataset's experimental results demonstrated the proposed transformer-MIL's efficiency and attention-based knowledge distillation. The noise-reduction-based attention cross-fusion network model (NRCA-FCFL) integrates multi-scale image information and generates a diagnostic model by feature fusion through a cross-attention mechanism, and the model is verified to outperform the state-of-the-art in a wide range of experiments [21]. Several experiments demonstrated that hybrid CNN-Transformer networks to integrate global and local information could yield better results [22]. The GasHis-Transformer model integrates the descriptive power of ViT and CNN for both global and local information. It obtained good classification performance on gastric histopathology images and showed excellent generalization ability on other histopathology image datasets [23]. Therefore, the transformer model based on the attention mechanism will further promote the innovation of histopathology image analysis techniques.

However, based-CNNs require a large amount of data and relevant annotations when training diagnostic models. Second, deep learning models are not very interpretable, and the hardware equipment required for training models is high. These drawbacks bring great resistance to the application of based-CNNs. The attention-based transformer model complements the convolution, has stronger modelling capabilities, is scalable on large data, and can better link language and vision. Compared with based-CNNs, advantages such as fewer parameters, faster training, and better diagnosis will be widely used in histopathological image analysis. Therefore, this paper proposes a Swim framework based on local deep convolution for the multi-label classification of organized histopathological images to comprehensively analyze the microenvironment of liver tumors. The main contributions of this paper are as follows:

1) This paper proposes a locally deep convolutional Swim framework to classify multi-label histopathological images to obtain local visual field diagnosis results. The tumor-to-stroma ratio can be calculated from the multi-label classification results of small slices in the WSI.

2) The LDC module in the LDCSF uses multi-channel information feedforward to extract detailed features from the Swin transformer to obtain a low-dimensional feature matrix. The FR module is joined by skipping, and uses the corresponding weight factor vector obtained from each channel to do point multiplication with the original feature vector matrix to generate a representative feature map.

3) Through extensive number of experiments, we compared the classification performance of DeiT, T2T-ViT, PVT, TNT, Swin transformer and CrossViT models on multi-label data sets, and proved the deep convolution module and FR module through ablation experiments. network, which provides a baseline for future classification work.

## II. MATERIALS

### A. Dataset

The liver tumor whole-slide images (WSIs) used in the experiment were obtained by clinical doctors from five liver cancer patients at the Third Xiangya Hospital [24]. A total of five WSIs were collected. Subsequently, pathologists from the Department of Pathology at Xiangya Hospital outlined the non-tumor, tumor, necrosis and Interstitial area based on the WSIs. For the multi-label classification task, each of the five WSIs was divided into smaller patches of size 224 × 224, resulting in a total of 68,175 patches. Since some patches overlapped between two regions, these overlapping patches were labeled with two corresponding labels. In addition, to avoid class imbalance in the dataset due to the smaller number of necrotic patches, the proportion of patches for each category was controlled within 1:3, preventing overfitting of the diagnostic model during training. Finally, patches of different types were organized into a multi-label dataset. Detailed information about the labels and quantities in the dataset is provided in TABLE I.

TABLE I INUMBER OF SAMPLES WITH DIFFERENT LABELS IN THE MULTI-LABEL DATASET.

| Label | Train dataset | Test dataset |
| --- | --- | --- |
| Interstitial area | 3400 | 12839 |
| Interstitial area & Non-Tumor | 2946 | 1490 |
| Interstitial area & Tumor | 3300 | 3279 |
| Necrosis | 1436 | 669 |
| Non-Tumor | 3000 | 9557 |
| Tumor | 2300 | 2010 |

Tissues are usually composed of cells, and different tissues exhibit different cellular characteristics. Images observed under high magnification microscopy capture information about the shape of the cells, but under low magnification can capture information about the structure of the cells. Histopathological images of a patient with liver cancer are depicted in Fig. 1. The Interstitial area shows a sparse distribution with some nuclei scattered in the intercellular fluid; the nuclei in Non-Tumor are neatly arranged with tight connections between cells and cells; the nuclei in Tumor are larger and unevenly arranged with loose connections between cells and cells and adhesions. The nuclei in Tumor are larger and unevenly arranged, with loose connections between cells and cells and low adhesion capacity; Necrosis does not contain any nuclei or organelles. Because cancer tissue has both cellular and structural heterogeneity, images taken at multiple magnifications will contain important information. Pathologists diagnose disease by varying the magnification of the microscope to obtain different types of information from the cellular level to the tissue level.

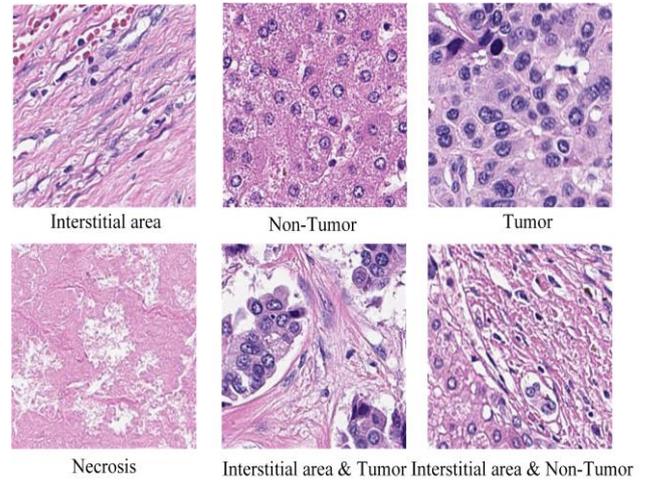

Fig. 1. Characteristics of different samples in the multi-label dataset.

## III. METHOD

This section describes our proposed LDCSF for multi-label classification of histopathological images. As shown in Fig. 2, clinical pathologists generate WSIs, which are ultra-high-resolution images with a size of $25000 \times 25000$, by removing pathological data from glass slides using an electron microscope. WSIs are then divided into numerous small patches to construct a multi-label image dataset for LDCSF learning. The main task of LDCSF is to provide multi-label diagnostic results within the local view, and it consists of three components: the Swin Transformer module, the Local Deep Convolution (LDC) module, and the Feature Reconstruction (FR) module.

### A. Swin transformer

The Swin transformer is a generalized attention-based mechanism for computer vision backbone networks. It is widely used in various granular-level classification tasks or segmentation tasks, and its robust performance has been verified in region target detection, pixel-level semantic segmentation, image-level image classification, etc. The core idea of the Swin transformer is to combine the highly modelable transformer structure with important a priori knowledge of visual signals, including hierarchy, localization and equilibrium invariance, etc.

As shown in Fig. 2, the feature extraction and classification of images are done in four main steps. Among them, stage 1, 2 and 3 are composed of the Swin transformer module, LDC module and FR module. First, the pathological image patches of RGB are first generated with non-overlapping patches by the patch segmentation module. Each patch is considered a token, and its features are set as a splice of the original pixel RGB. In the experiments for the classification task, our patch size is $4 \times 4$, and the feature dimension of each patch is $4 \times 4 \times 3 = 48$. The number of patches is $H/4 \times W/4$; all tokens are projected to arbitrary dimensions (denoted as $C$) by the convolutional computation of the linear embedding layer.

In the stage1 part, the input is divided into patches of the same size through linear embedding, the feature dimension becomes $C$, and then sent to the Swin Transformer Block, the LDC module further learns the features, and the multi-channel feature map is passed to the FR module to generate a new feature map. As the network layers deepen, the patch merging layer between stage1 and stage2 is used to reduce the number of tokens. The first patch merging layer splices the adjacent features of each group of $2\times 2$ patches, and the number of patches blocks becomes $H/8\times W/8$, and sends the spliced $4C$-dimensional features to the linear layer for processing to generate $2C$-dimensional features [25]. Stage 2, stage 3, and stage 4 parts are similar to the main structure of the stage1 part, and The number of patches processed is $H/8\times W/8$, $H/16\times W/16$, $H/32\times W/32$, and the dimensions of these patches in different Stages are $2C$, $4C$, $8C$, respectively. Stage 5 is a residual neural network undertaking the classification function [14].

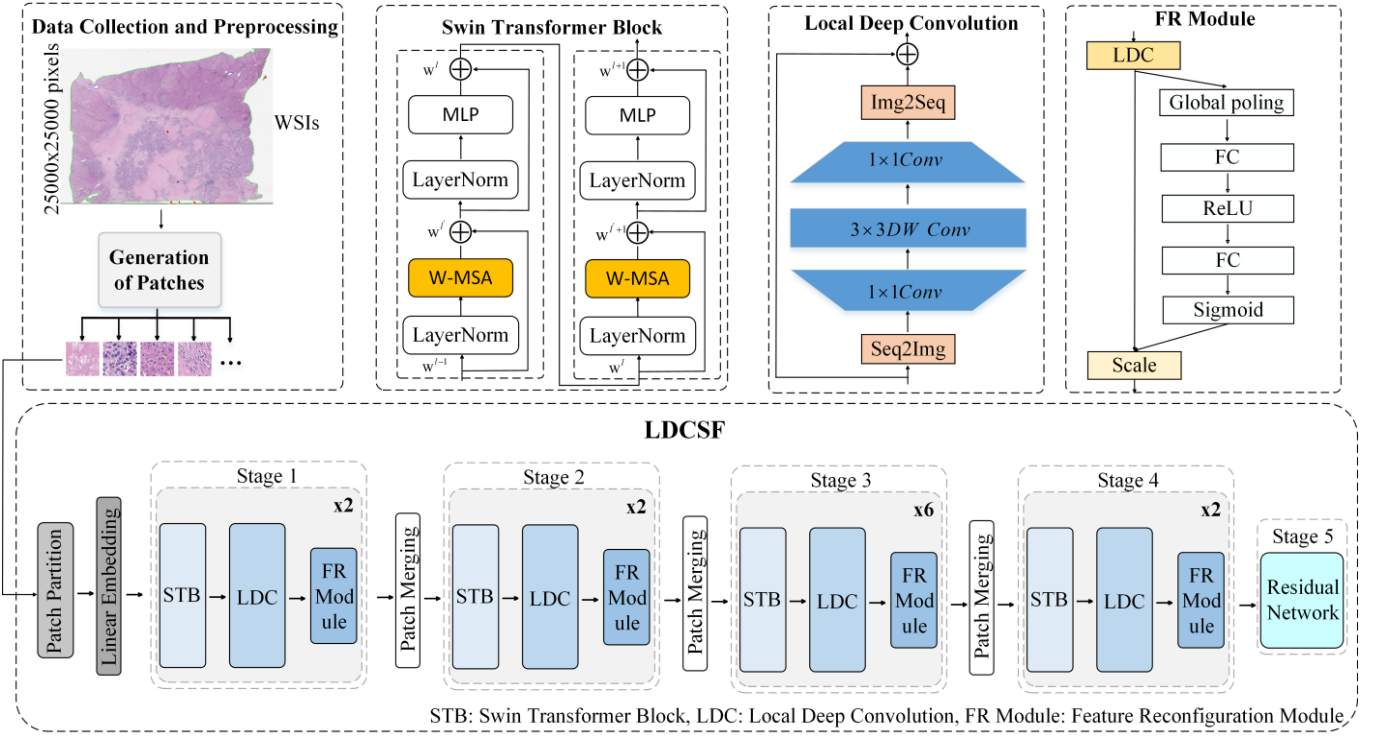

Fig. 2. Flow chart of LDCSF processing multi-label classification WSI, including data collection and preprocessing, block diagram of Swim Transformer block, Local Deep Con and FR model.

The Swin transformer block consists of a standard multi-headed self-attentive (MSA) module based on shifted windows, followed by a two layers MLP with GELU nonlinearity in the middle. An important design feature of the Swin transformer is the shifted windows, which significantly reduces the complexity of the algorithm and allows it to grow linearly with the size of the input; at the same time, due to the different sliding windows, the design of the non-overlapping windows is more hardware-friendly and thus has a faster real-world speed. Assume that the computational complexity of a window based on $h\times w$ patched images is:

$$\phi(W-MSA) = 4hwC^2 + 2M^2hwC \quad (1)$$

Where the former is quadratic to patch number $hw$, and the latter is linear when $M$ is fixed (set to 7 by default). To address the impact of the lack of cross-window connectivity problem, the shift-window partitioning method alternates among neighbouring the Swin transformer blocks. In the self-attention mechanism, we introduce the existing method to compute similarity with each head containing the relative position deviation B [26]:

$$Attention(Q,K,V) = soft\max(QK^T\sqrt{d}+B)V, \quad (2)$$

Where $Q,K,V \in R^{M^2+d}$ are the query, key and value matrices; $d$ is the query/key dimension, , and $M^2$ is the number of patches in a window.

B. Local deep convolution

However, since the attention mechanism of the Swin transformer only captures the global dependencies between tokens. The Swin transformer does not model the analysis of local dependencies between adjacent pixels. LDC is an effective way to introduce locality into the network. As shown in Fig. 2, "DW" means depth convolution. In order to handle the convolution operation, the conversion between the sequence and the image feature map is added through "Seq2Img" and "Img2Seq" in the FR module. With this in mind, we reintroduce depth convolution in the transformer feedforward network. The LDC module is composed of a convolutional layer, a batch normalization layer and an H_Swish activation layer. 8C-dimensional token sequences are first reshaped in a feature mapping rearranged on a 2D lattice and then fed into a 2D convolutional layer. $k\times k(k>1)$ convolutional kernels aggregate the features in multiple channels, compute a new feature to learn a richer feature representation and pass the generated feature map to the next layer. The computation can be expressed as:

$$Y^r = f(f(Z^r \otimes W_1^r) \otimes W_d) \otimes W_2^r \qquad (3)$$

Where $W_d \in R^{r^{d \times 1 \times k \times k}}$ is the kernel of the depth-wise convolution. Numerous experiments have demonstrated that the H_Swish activation function can be used to expand the capacity of the network and improve its efficiency of the network. Therefore, the H_Swish activation layer is used as a part of the LDC module.

*C. Feature Reconfiguration Module*

The feature map after LDC is passed to the FR module for critical operations. First, the squeeze operation, which performs feature compression along the spatial dimension, turns each two-dimensional feature channel into a real number that somehow has a global perceptual field and matches the output dimension to the number of input feature channels. It characterizes the global distribution of the response over the feature channels and makes the global perceptual field also available for layers close to the input, which can be expressed as:

$$Z_c = F_{sq}(U_c) = \frac{1}{H \times W} \sum_{i=1}^{H} \sum_{j=1}^{W} u_c(i,j) \qquad (4)$$

The main focus is to convert the $H \times W \times C$ features into a $1 \times 1 \times 1$ output. Next is the excitation operation, which is a mechanism similar to gates in recurrent neural networks. Weights are generated for each feature channel by means of a parameter $W$, where the parameter $W$ is learned to explicitly model the correlation between feature channels. Finally, a reweight operation treats the weights of the excitation output as the significant features of each feature channel after feature selection and then re-calibrates the original features in the channel dimension by multiplying the weights channel by channel onto the previous features.

*D. Classification module*

The residual neural network serves as the classification module of the whole classification network, and its residual connections can effectively solve the problem of network degradation and accelerate the convergence of the results so that the classification results are faster. Second, the residual network is also designed to prevent overfitting. The role of randomly discarding some neurons solves the overfitting problem that occurs in the case of small data sets and large models to some extent. The experiments were designed with a loss according to each category to help the model independently and dynamically adjust the error for each label, which can be expressed as

$$L = l_i + l_{nt} + l_t + l_n \qquad (5)$$

L is the total loss of the model, $l_i$ is the loss of interstitial area; $l_{nt}$ is the loss of non-tumor; $l_t$ is the loss of tumor; $l_n$ is the loss of necrosis. All the losses are calculated using the cross-entropy loss function.

## IV. EXPERIMENTS

*A. Experiment steps*

In order to avoid the classification error of the model on different labels due to the unbalanced sample size, for all training models, data augmentations such as horizonal and vertical flipping are used to increase data diversity and randomHSV is also adopted to randomly change the hue, saturation, and value of images in the hue-saturation-value (HSV)color space, making the model robust to color perturbations. In the classification task, we first train DeiT, T2T-ViT, PVT, TNT, Swin transformer and CrossViT. Then, we add the LDC module and FR module (LF) to PVT, TNT, T2T and CrossViT respectively to construct new networks to compare with the LDCSF.

*B. Implementation details*

Our experiments are achieved based on Python 3.8 and Pytorch 1.11.0. We train our model on a Nvidia GeForce RTX 4090 GPU with 24GB memory. The input image size and max epoch are set as $224 \times 224$ and 150, respectively. During the training period, the default batch size is 24 and the SGD optimizer with momentum 0.9 and weight decay 1e-4 is used to optimize our models for back-propagation. The model is trained with the learning rate set to 0.001. In our experiments, we use random cross-validation to train our model. We randomly divide the training, validation, and test data 10 times, then train and test each model, and its final results are averaged.

*C. Evaluation*

In order to understand the generalization ability of the model, we need to measure it by some metric. For the classification task, the experiments chose precision, recall, F1-score, accuracy, confusion matrix and ROC curve to evaluate the model's performance comprehensively. Precision represents the percentage of correct positive samples to predicted positive samples; recall represents the percentage of correct positive samples to predicted correct samples; F1-score is a combined evaluation of precision and recall; and accuracy represents the percentage of correct samples to predicted samples [27]. The confusion matrix represents the probability of correct and incorrect for each label. The true positive rate (TPR) and false positive rate (FPR) in the ROC curve reflect the predictive power of the model [28]. Since LDCSF are multi-label classification networks, we will carefully analyze and discuss the LDCSF network on each label. Finally, in the segmentation task, experiments were conducted to assess the generalization ability of the model using dice-similarity coefficient (DICE) and accuracy. Dice is the most frequently used metric in medical segmentation tasks and is an ensemble similarity measure that is usually used to calculate the similarity of two samples, the closer to 1 the better the effect [29].

## V. RESULTS

*A. Comparison with state-of-the-art models*

TABLE II statistically shows the values of precision, recall, F1-score and accuracy of interstitial area obtained from 11 classification networks tested on the dataset. The Swin ViT achieves 0.9515, 0.9471, and 0.9487 for recall, F1-score, and accuracy, respectively, which exceeds the performance of all other classification networks. Swin+LF achieves 0.9443 for precision, which outperforms other classification networks. Therefore, the Swin transformer block model plays an important role in improving the classification accuracy, sensitivity and specificity in the interstitial area.

TABLE II  THE PREDICTION RESULTS OF LABEL INTERSTITIAL AREA ARE OBTAINED BY 13 CLASSIFICATION NETWORKS ON THE TEST SET.

| Model | Precision | Recall | F1-score | Accuracy |
|---|---|---|---|---|
| DeiT | 0.7131 | 0.5992 | 0.5751 | 0.6697 |
| TNT | 0.9376 | 0.9428 | 0.9399 | 0.942 |
| T2T | 0.6675 | 0.5827 | 0.5572 | 0.652 |
| Swin ViT | 0.9438 | **0.9515** | **0.9471** | **0.9487** |
| CrossViT | 0.9412 | 0.9291 | 0.9343 | 0.9377 |
| PVT+LF | 0.8668 | 0.8405 | 0.8493 | 0.8599 |
| TNT+LF | 0.9257 | 0.9344 | 0.9292 | 0.9313 |
| T2T+LF | 0.9436 | 0.9454 | 0.9445 | 0.9466 |
| CrossViT+LF | 0.9427 | 0.9374 | 0.9399 | 0.9426 |
| Swin+LF | **0.9443** | 0.9484 | 0.9462 | 0.9481 |
| LDCSF | 0.9408 | 0.9491 | 0.9443 | 0.946 |

TABLE III counts the precision, recall, F1-score and accuracy values of interstitial area obtained by 11 classification networks tested on the dataset. CrossViT achieves 0.9895 in precision, probably due to the learning of cross attention, which improves the model's classification performance. LDCSF reached 0.9963 on recall; Swin ViT reached 0.9896 on F1-score; T2T+LF reached 0.9973 on accuracy. Because there are a large number of the Swin transformer blocks in LDCSF and Swin ViT, which can help the model improve performance. The classification effect of T2T ranks last among all models. However, the performance of T2T+LF is better. It may be that the LS module plays a key role in helping the T2T model to obtain key feature information in feature extraction.

TABLE III  THE PREDICTION RESULTS OF LABEL NECROSIS ARE OBTAINED BY 13 CLASSIFICATION NETWORKS ON THE TEST SET.

| Model | Precision | Recall | F1-score | Accuracy |
|---|---|---|---|---|
| DeiT | 0.9614 | 0.691 | 0.804 | 0.9704 |
| TNT | 0.9512 | 0.9863 | 0.9679 | 0.9893 |
| T2T | 0.8505 | 0.9615 | 0.8959 | 0.9612 |
| Swin ViT | 0.9844 | 0.995 | **0.9896** | 0.9966 |
| CrossViT | **0.9895** | 0.9663 | 0.9776 | 0.993 |
| PVT+LF | 0.9389 | 0.6209 | 0.6754 | 0.9322 |
| TNT+LF | 0.9455 | 0.9856 | 0.9645 | 0.9881 |
| T2T+LF | 0.9907 | 0.9922 | 0.9914 | **0.9973** |
| CrossViT+LF | 0.9814 | 0.9844 | 0.9829 | 0.9945 |
| Swin+LF | 0.9788 | 0.9893 | 0.984 | 0.9948 |
| LDCSF | 0.9798 | **0.9963** | 0.9878 | 0.996 |

Table IV counts the precision, recall, F1-score and accuracy values of non-tumor obtained by 11 classification networks tested on the dataset. LDCSF outperforms the other 10 classification networks by 0.9776, 0.9807, 0.9791, and 0.9808 in precision, recall, F1-score and accuracy, respectively. First, this is probably because the shifted non-overlapping windows in the Swin transformer block are helpful for feature extraction, and the LDC module pays more attention to the relationship between adjacent pixels and transfers the features to FR module in parallel, multi-channel. The FR module extracts the effective information in the feature map again and combines it with the original map to generate a new feature map. Second, the prediction accuracy and sensitivity of DeiT, TNT, T2T, Swin ViT, and CrossViT are not high, probably because these models do not achieve the best results in feature extraction.

TABLE IV  THE PREDICTION RESULTS OF LABEL NON TUMOR ARE OBTAINED BY 13 CLASSIFICATION NETWORKS ON THE TEST SET.

| Model | Precision | Recall | F1-score | Accuracy |
|---|---|---|---|---|
| DeiT | 0.7447 | 0.74 | 0.7421 | 0.7656 |
| TNT | 0.9689 | 0.97 | 0.9694 | 0.9719 |
| T2T | 0.7901 | 0.8152 | 0.7909 | 0.797 |
| Swin ViT | 0.9641 | 0.9726 | 0.968 | 0.9704 |
| CrossViT | 0.9684 | 0.9677 | 0.968 | 0.9707 |
| PVT+LF | 0.4571 | 0.4849 | 0.4282 | 0.6041 |
| TNT+LF | 0.9735 | 0.9646 | 0.9688 | 0.9716 |
| T2T+LF | 0.9747 | 0.9784 | 0.9765 | 0.9783 |
| CrossViT+LF | 0.9622 | 0.9704 | 0.966 | 0.9686 |
| Swin+LF | 0.9665 | 0.9635 | 0.965 | 0.9679 |
| LDCSF | **0.9776** | **0.9807** | **0.9791** | **0.9808** |

TABLE V counts the values of precision, recall, F1-score and accuracy obtained for Tumors for 11 classification networks tested on the dataset. CrossViT+LF achieves 0.9827 on precision. however, CrossViT+LF is slightly higher than CrossViT on precision, recall, F1-score and accuracy. This is because the LS module plays a key role in feature extraction. LDCSF achieves 0.9838, 09832,0.9847 for recall, F1-score, and accuracy, respectively. block and LS module make LDCSF achieve the best performance of classification network.

TABLE V  THE PREDICTION RESULTS OF LABEL TUMOR ARE OBTAINED BY 13 CLASSIFICATION NETWORKS ON THE TEST SET.

| Model | Precision | Recall | F1-score | Accuracy |
|---|---|---|---|---|
| DeiT | 0.874 | 0.8673 | 0.8705 | 0.8837 |
| TNT | 0.9817 | 0.9805 | 0.9811 | 0.9829 |
| T2T | 0.8641 | 0.6735 | 0.6839 | 0.7729 |
| Swin ViT | 0.9795 | 0.9793 | 0.9794 | 0.9814 |
| CrossViT | 0.9788 | 0.978 | 0.9784 | 0.9805 |
| PVT+LF | 0.79 | 0.814 | 0.7614 | 0.7631 |
| TNT+LF | 0.9657 | 0.9783 | 0.9714 | 0.9737 |
| T2T+LF | 0.9815 | 0.9829 | 0.9822 | 0.9838 |
| CrossViT+LF | **0.9827** | 0.9795 | 0.9811 | 0.9829 |
| Swin+LF | 0.9771 | 0.9837 | 0.9803 | 0.982 |
| LDCSF | 0.9826 | **0.9838** | **0.9832** | **0.9847** |

The confusion matrix is also a key metric to evaluate the generalizability of the model. 11 models were validated on the test set to obtain the confusion matrix for each label.

The ROC curve is used to explain the performance of the 11 classification networks. The steeper the curve, the better the performance of the classification network. We plotted 6 ROC curves for each network, including micro-average, macro-average, interstitial area, necrosis, non-tumor and tumor. The ROC curve below the diagonal indicates a poor method for classifying networks. It can be seen from Fig. 3 that the PVT+LF model has a large error on the non-tumor. DeiT and T2T do not perform well on most labels. The micro-averaged ROCs of TNT, CrossViT, TNT+LF and CrossViT+LF were all between 0.95 and 0.97. The micro-average ROC of T2T, Swin ViT, T2T+LF, Swin+LF and LDCSF are above 0.97. Among them, LDCSF has reached the highest level in history. On different labels, the ROC curve of interstitial area is not as steep as that of other labels, probably because the diversity of interstitial area causes the model to generalize worse than other labels. In summary, we can conclude that the network with the LS module added is better than the network without the LS module in classification performance. Therefore, our designed LDCSF can serve as a standard method to classify liver cancer WSI.

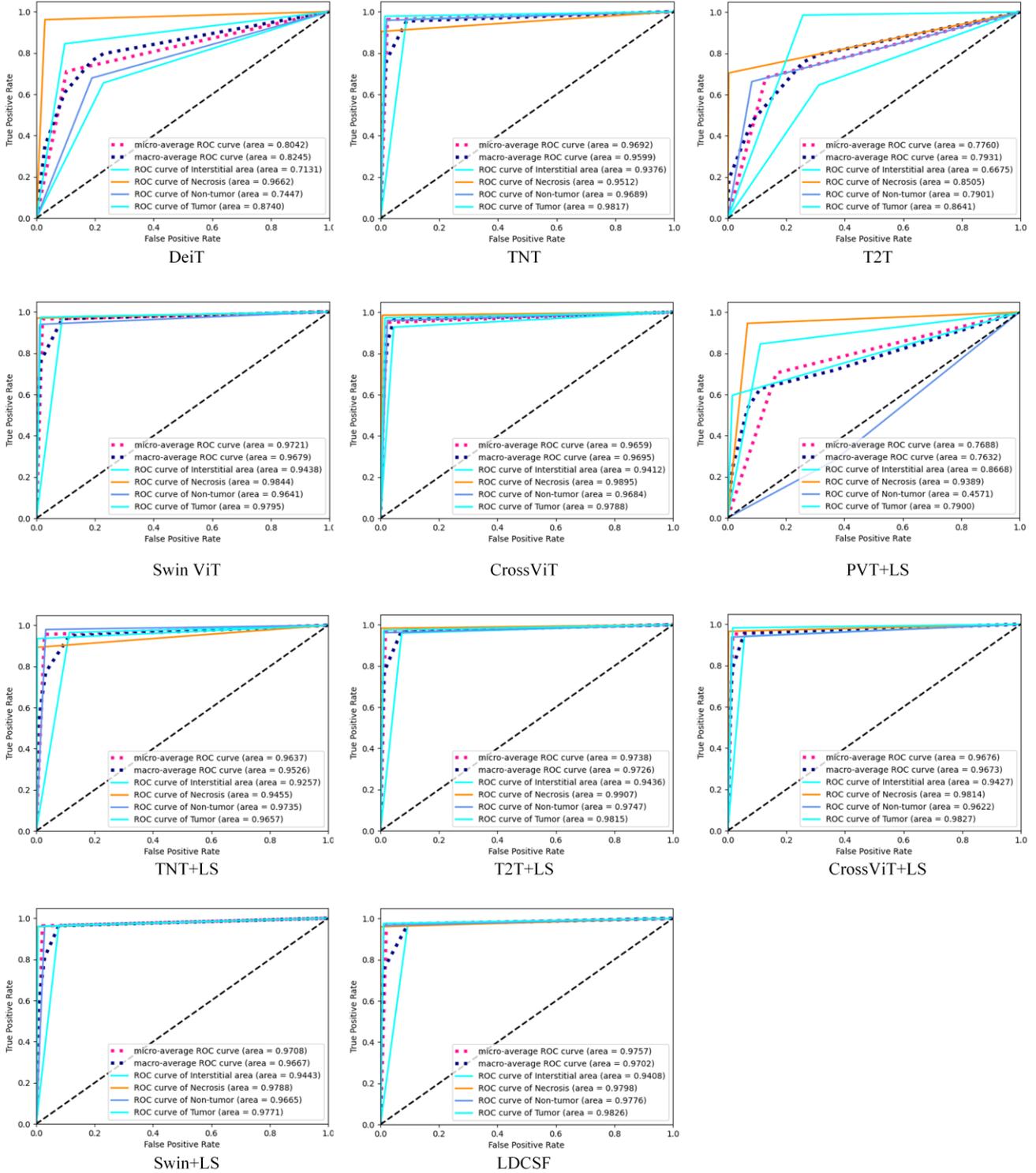

Fig. 3. ROC curves plotted on the test dataset with 11 classification networks.

*B. Ablation experiments*

**Impact of LDC module**: To evaluate the impact of the deep convolution module on the LDCSF, we removed the

LDC module from the LDCSF, trained a new model, and obtained prediction results with the new model. As shown in TABLE VI, the accuracy obtained by the new model with the LDC module removed exceeds that of the LDCSF only on the interstitial area. The accuracy of the LDCSF predictions on necrosis, non-tumor and tumor exceeds that of the new model by 0.0015, 0.0049, and 0.0024, respectively. This may be because the LDC module plays an important feature extraction role. Although the LDCSF model prediction accuracy decreases for interstitial area, the model prediction accuracy increases for all other labels. Therefore, we consider the LDC module as the core component of the LDCSF network.

TABLE VI    IMPACT OF THE LDC MODULE ON THE LDCSF NETWORK EVALUATED BY ACCURACY.

|     | Interstitial area | Necrosis | Non-tumor | Tumor |
|-----|-------------------|----------|-----------|-------|
| NO  | 0.9518            | 0.9945   | 0.9759    | 0.9823 |
| YES | 0.946             | 0.996    | 0.9808    | 0.9847 |

**Impact of FR module**: To evaluate the impact of the FR module on the LDCSF, we removed the FR module from the LDCSF, trained a new model, and obtained prediction results with the new model. As shown in TABLE VII, the accuracy obtained by the new model with the FR module removed exceeds that of the LDCSF only on the interstitial area. The accuracy of the LDCSF network predictions on necrosis, non-tumor and tumor exceeds that of the new model by 0.0006, 0.0009, and 0.0006, respectively. This may be because the FR module plays a feature extraction plays an important role. Although the LDCSF prediction accuracy decreases for interstitial area, the model prediction accuracy increases for all other labels. Therefore, we consider the FR module as the core component of the LDCSF.

TABLE VII    IMPACT OF THE FR MODULE ON THE LDCSF NETWORK ASSESSED BY ACCURACY.

|         | Interstitial area | Necrosis | Non-tumor | Tumor |
|---------|-------------------|----------|-----------|-------|
| Without | 0.9509            | 0.9954   | 0.9799    | 0.9841 |
| with    | 0.946             | 0.996    | 0.9808    | 0.9847 |

## VI. CONCLUSION

This paper proposes a LDCSF for the multi-label classification of organized histopathological images. LDCSF consists mainly of the Swin Transformer, LDC, and FR modules. The Swin Transformer utilizes self-attention sliding windows to interact with information between adjacent positions in the image. LDC performs two-dimensional reconstruction of features, efficiently transferring the features to the FR module through multi-channel convolution calculations. The FR module performs dot product operations using the corresponding weight coefficient vectors obtained from the channels and the original feature map vectors, generating representative feature maps. On the test set, LDCSF achieves classification accuracies of 0.9460, 0.9960, 0.9808, and 0.9847 for the stromal region, necrosis, non-tumor, and tumor, respectively. The experimental results demonstrate that LDCSF can obtain multi-label classification results for local regions. By analyzing the distribution areas and quantities of different labels in WSIs, it can further calculate the tumor stroma ratio. Finally, LDCSF lays the foundation for preliminary analysis of the tumor microenvironment in liver cancer.

ACKNOWLEDGMENT

This work was supported by National Key R&D Program of China 2022YFC3400400; NSFC Grants U19A2067; Top 10 Technical Key Project in Hunan Province 2023GK1010, Key Technologies R&D Program of Guangdong Province (2023B1111030004 to FFH). The Funds of State Key Laboratory of Chemo/Biosensing and Chemometrics, the National Supercomputing Center in Changsha (http://nscc.hnu.edu.cn/), and Peng Cheng Lab.